\documentclass[conference]{IEEEtran}
\IEEEoverridecommandlockouts
\usepackage{cite}
\usepackage{amsmath,amssymb,amsfonts}
\usepackage{algorithmic}
\usepackage{graphicx}
\usepackage{hyperref}
\usepackage{textcomp}
\usepackage{xcolor}
\usepackage{listings}
\usepackage{pythonhighlight}
\usepackage{url}
\usepackage[normalem]{ulem}
\def\BibTeX{{\rm B\kern-.05em{\sc i\kern-.025em b}\kern-.08em
    T\kern-.1667em\lower.7ex\hbox{E}\kern-.125emX}}
\begin{document}
\title{PyMilo: A Python Library for ML I/O}

\author{
    \IEEEauthorblockN{
        AmirHosein Rostami\IEEEauthorrefmark{1}\IEEEauthorrefmark{2}, Sepand Haghighi\IEEEauthorrefmark{1}, Sadra Sabouri\IEEEauthorrefmark{1}\IEEEauthorrefmark{3}, Alireza Zolanvari\IEEEauthorrefmark{1}\IEEEauthorrefmark{4},
    }
    \IEEEauthorblockA{\IEEEauthorrefmark{1} Open Science Lab\\
    \{amirhosein, sepand, sadra, alireza\}@openscilab.com}
    \IEEEauthorblockA{\IEEEauthorrefmark{2} University of Toronto}
    \IEEEauthorblockA{\IEEEauthorrefmark{3} University of Southern California}
    \IEEEauthorblockA{\IEEEauthorrefmark{4} University of Groningen}
}

\maketitle

\begin{abstract}
PyMilo is an open-source Python package that addresses the limitations of existing Machine Learning (ML) model storage formats by providing a transparent, reliable, and safe method for exporting and deploying trained models. Current formats, such as pickle and other binary formats, have significant problems, such as reliability, safety, and transparency issues. In contrast, PyMilo serializes ML models in a transparent non-executable format, enabling straightforward and safe model exchange, while also facilitating the deserialization and deployment of exported models in production environments. This package aims to provide a seamless, end-to-end solution for the exportation and importation of pre-trained ML models, which simplifies the model development and deployment pipeline.
\end{abstract}

% we say tranparent, reliable and safe and then we say that other tools don't have this, reliability, safety and transparency issues.

% TODO performance degradation


\section{Introduction}
The Python programming language has emerged as a top choice for machine learning applications, owing to its simplicity, flexibility, and vast community support~\cite{raschka2020machine}. However, despite the numerous tools and libraries developed by the Python community, a significant challenge persists: sharing and deploying trained machine learning models in a secure and transparent manner. This concern is particularly pressing in high-stakes domains such as healthcare, where model integrity and accountability are essential~\cite{garbin2022assessing}. Lack of transparency and safety in model sharing can have severe consequences, including the potential for malicious code injection and data breaches. \\
The widespread use of binary formats for saving machine learning models~\cite{raschka2015python, raschka2019python, brownlee2020save}, such as pickle and joblib~\cite{brownlee2020save}, has introduced significant security risks, as these formats are not human-readable and can be exploited by malicious actors~\cite{verma2023insecure}. 
Furthermore, tools such as ONNX and PMML convert the original model into alternative representations with structural differences. This conversion enhances compatibility across diverse environments, but it also introduces new challenges, such as performance degradation.

\begin{figure*}[ht]
    \centering
    \includegraphics[width=0.8\textwidth]{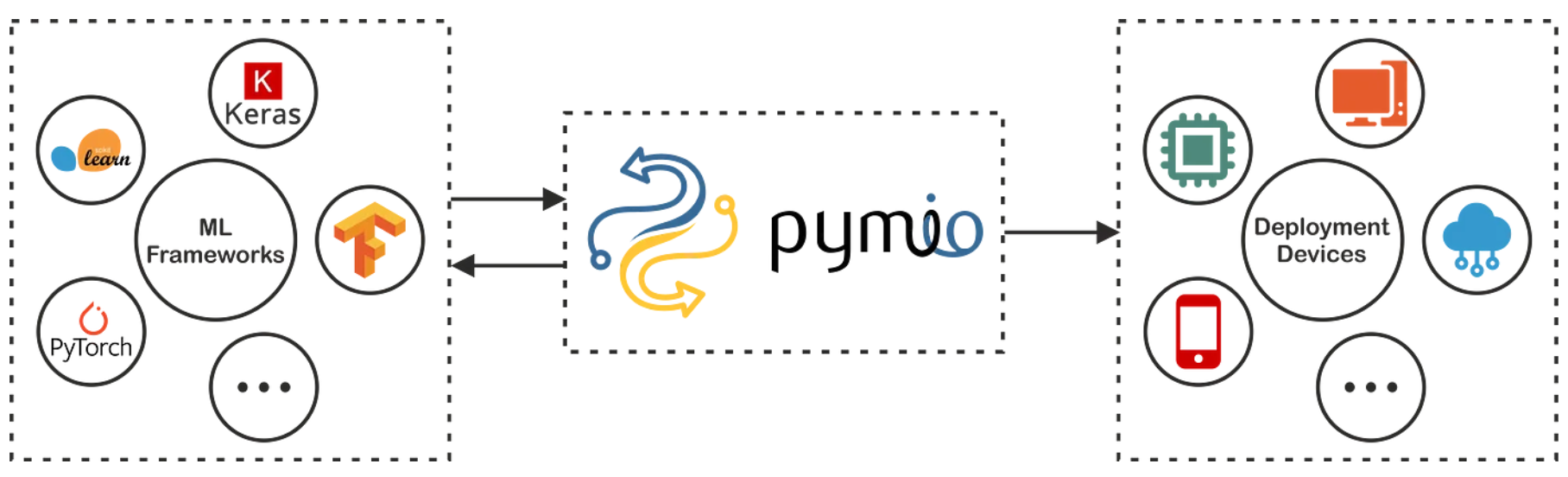}
    \caption{PyMilo is an end-to-end, transparent, and safe solution for transporting machine learning models from machine learning frameworks to target devices. Unlike other tools that transform models into alternative representations with structural differences, PyMilo preserves the original model's structure, allowing it to be imported back as the exact same object in its native framework.
    }
    \label{fig:PyMilo}
\end{figure*}

To address these concerns, the machine learning community requires an end-to-end transparent way of communicating and verifying model information. This necessitates the ability to inspect model parameters, which involves machine learning models and their components to be transparently serializable. \\
Here in this paper, we introduce a transparent, non-executable, safe and human-readable approach for serializing and deserializing machine learning models. Our approach is implemented as a Python package named PyMilo\footnote{\url{https://github.com/openscilab/pymilo}}. PyMilo serves as a transporter that serializes trained models generated from machine learning frameworks into a transparent, nonexecutable, safe and human-readable  formats, like JSON, and deserializes it back into the exact same original machine learning model (Figure \ref{fig:PyMilo}). PyMilo transportation is completely end-to-end, providing a standalone process from exporting to importing machine learning models. ``End-to-end'' here refers to a full process from exporting the trained model to importing and retrieving the same original machine learning model. This approach enables the transportation of machine learning models to any target device, where they can be executed in inference mode without requiring additional dependencies. This provides a general solution for creating human-readable, transparent, and safe machine learning models that can be easily shared, inspected, and deployed. PyMilo benefits a wide range of stakeholders, including machine learning engineers, data scientists, and AI practitioners, by facilitating the development of more transparent and accountable AI systems. Furthermore, researchers working on transparent AI~\cite{rauker2023toward}, user privacy in ML~\cite{bodimani2024assessing}, and safe AI~\cite{macrae2019governing} can use PyMilo as a framework that preserves transparency and safety in a machine learning environment. In the next section, we will review existing solutions and related efforts in serialization of machine learning models to provide a better understanding of the context and relevance of our approach.
\subsubsection*{Related works}
In this section, we review existing approaches for transporting (serializing \& deserializing) trained machine learning models in Python, highlighting their limitations and security concerns.\\
\textbf{Pickle} is the standard Python library to serialize and deserialize Python objects into binary files. While it provides a convenient way to store and load Python objects, the use of Pickle is not secure due to the risk of executing malicious code during deserialization. As warned in the Python documentation\footnote{\url{https://docs.python.org/3/library/pickle.html}}, Pickle is not intended to be secure against erroneous or maliciously constructed data. This vulnerability can lead to arbitrary code execution, making it a potential security risk. As an alternative, the use of JSON serializers is recommended whenever possible.\\
\textbf{SKOPS}~\cite{jalali2022skops} is a Python library that facilitates sharing and deploying scikit-learn-based models. Although it provides tools for integrating models with the Hugging Face Hub, its binary output format still has the risk of malicious code injection. Moreover, SKOPS is limited to scikit-learn models and does not support other popular machine learning frameworks.\\
\textbf{TensorFlow.js}~\cite{yu2018tfjs} is an open-source JavaScript library to train and deploy TensorFlow models. Although it provides hardware-accelerated support, it is limited to TensorFlow models and can be inefficient when dealing with large models~\cite{tensorflow2015-whitepaper}. Additionally, TensorFlow.js requires significant modifications to the original model architecture, which can lead to compatibility issues and performance degradation.\\
\textbf{ONNX}~\cite{chen2017onnx} is an open format to represent machine learning models. Its approach involves reconstructing models using a set of predefined building blocks that does not preserve the original model's structure and behavior. However, this reconstruction process can lead to accuracy losses and compatibility issues. For example, ONNX has shown a significant performance degradation during model export, with \textbf{up to 10-15\% accuracy loss} in certain scenarios \cite{onnx_performance2020}. Furthermore, ONNX lacks transparency in its output.

In contrast with the above-mentioned tools, PyMilo provides a transparent non-executable safe file format to serialize trained machine learning models in Python. By wrapping asymmetric cryptography around the transparent non-executable exported file, we both verify the authenticity of the publisher and also ensure that there are no security threats inside the given file. Having both of these features simultaneously is crucial for any collaborations that involve ML model sharing. This is especially important in sensitive domains like healthcare, where preserving patient data is a priority. Additionally, PyMilo plans to support multiple machine learning frameworks and provides a flexible and extensible architecture for future framework support. A general comparison is provided in table \ref{tab:PyMilo-compare}.

\begin{table}[t]
    \centering
    \caption{A comparison between PyMilo and other similar packages.}
    \resizebox{\linewidth}{!}{
    \begin{tabular}{c|c|c|c|c}
         Package & Transparent & Multi Framework & End-to-End & Secure \\ \hline

         Pickle & \textcolor{purple}{No} & \textcolor{teal}{Yes} & \textcolor{teal}{Yes} & \textcolor{purple}{No} \\ 
         SKOPS & \textcolor{purple}{No} & \textcolor{purple}{No} & \textcolor{teal}{Yes} & - \\
         Tensorflow.js & \textcolor{teal}{Yes} & \textcolor{purple}{No} & \textcolor{teal}{Yes} & \textcolor{teal}{Yes} \\
         ONNX & \textcolor{purple}{No} & \textcolor{teal}{Yes} & \textcolor{purple}{No} & - \\
         PyMilo & \textcolor{teal}{Yes} & \textcolor{teal}{Yes} & \textcolor{teal}{Yes} & \textcolor{teal}{Yes} \\ 
    \end{tabular}
    }
    \label{tab:PyMilo-compare}
\end{table}

This paper focuses on the features and plans for PyMilo version 1.1. The following sections are organized as follows. We first present the technical overview of PyMilo, highlighting its key components and design decisions. Then, we demonstrate the practical applicability of PyMilo through a sample use case, showcasing its functionality and benefits. Next, we describe the comprehensive testing system we developed to ensure the quality and reliability of the library. Finally, we discuss the limitations of our approach and outline future directions for the development of PyMilo.

\section{Technical Overview}
PyMilo facilitates the comprehensive export (serialization) of machine learning (ML) models, capturing their full algorithm, including structural architecture, parameter values, and custom functionality implementations. This ensures that models can be reconstructed precisely in new environments. As an end-to-end solution, it provides all the required tools for the precise reconstruction (deserialization) of the exported models. Additionally, PyMilo introduces a novel feature called ``ML Streaming" that simplifies the deployment and utilization of models in web environments. In the following, we present the architecture and capabilities of these two features.

\subsection{Model Serialization}
PyMilo's serialization mechanism captures the core structure and custom features of machine learning models. Each model has a unique structure that defines its behavior. For example, in decision trees, this includes attributes like the number of nodes and their connections. PyMilo ensures these essential components are stored, allowing accurate reconstruction of the model. Additionally, some models have custom components, such as specialized loss functions or optimization algorithms. PyMilo supports the serialization of these custom elements, ensuring models retain their original behavior across different platforms.

PyMilo employs the \emph{Chain of Responsibility} design pattern, introduced in~\cite{SV:2002}, to manage the serialization and deserialization of complex, non-serializable data structures. This design allows requests to be handled dynamically by multiple objects, fostering a flexible, extensible system. In PyMilo, this is achieved through a network of specialized Transporters, each responsible for serializing and deserializing specific data structures. The \verb|Transporter| interface defines the core methods, while an abstract class (\verb|AbstractTransporter|) provides partial implementations and establishes a blueprint for concrete Transporters. These concrete Transporters handle the specific serialization logic for various data types, ensuring that each unique data structure is efficiently processed according to its individual requirements. This architecture enables PyMilo to extend its support for additional data structures and serialization formats as needed, offering scalable and reliable data transport operations. Presented in Section~\ref{sec: functionallity} is an example of how PyMilo serializes and deserializes ``scikit-learn" models.

\begin{figure*}[!h]
    \centering
    \includegraphics[width=0.8\textwidth]{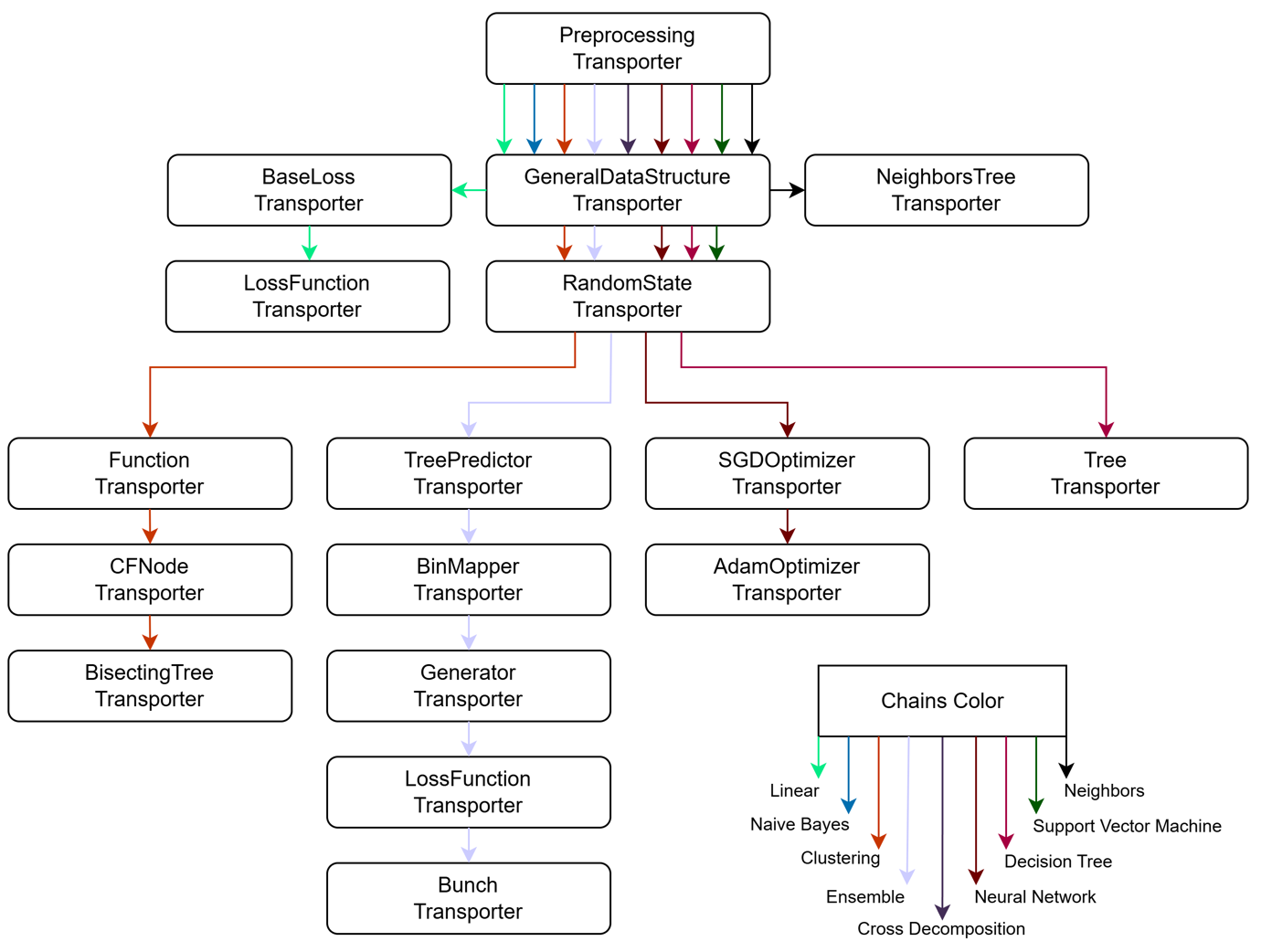}
    \caption{PyMilo's Transporter network is designed for scikit-learn based ML models. It uses the Chain of Responsibility pattern to construct specialized chains. Each chain has Transporters that manage the serialization and deserialization of specific data structures. Distinct chains are established for different machine learning model categories, such as decision trees and clustering.}
    \label{fig:scikit-trans}
\end{figure*}

From version 0.9, PyMilo supports all scikit-learn models through a custom implementation of 16 transporters designed to manage data structures that are used in the implementation of the scikit-learn models. Scikit-learn encompasses nine distinct categories of machine learning models: Linear Models, Neural Networks, Decision Trees, Clustering, Naive Bayes, Support Vector Machines (SVM), Nearest Neighbors, Cross Decomposition, and Ensemble Models. To facilitate the efficient transportation of each model category, PyMilo employs a chaining approach with different transporters, resulting in nine specialized chains tailored for each specific category. The complete arrangement of all 16 transporters and their interconnections is illustrated in Figure~\ref{fig:scikit-trans}.

\subsection{ML Streaming}
``ML Streaming" is designed to enhance PyMilo's functionality, allowing seamless deployment of exported models in web services to perform various tasks, such as streaming predictions, retraining models, and peer-to-peer model sharing. This feature introduces an efficient server-client architecture that facilitates real-time interaction with machine learning models. Using an instance of ``PyMilo Client", users forward incoming calls to an instance of ``PyMilo Server" hosting a serialized machine learning model to handle the tasks on their behalf. This architecture enables effortless interaction with complex models without requiring direct local storage or execution. ``ML Streaming" simplifies the interaction between distributed systems, offering a flexible framework for web-based machine learning tasks.

\begin{figure*}[!h]
    \centering
    \includegraphics[width=0.8\textwidth]{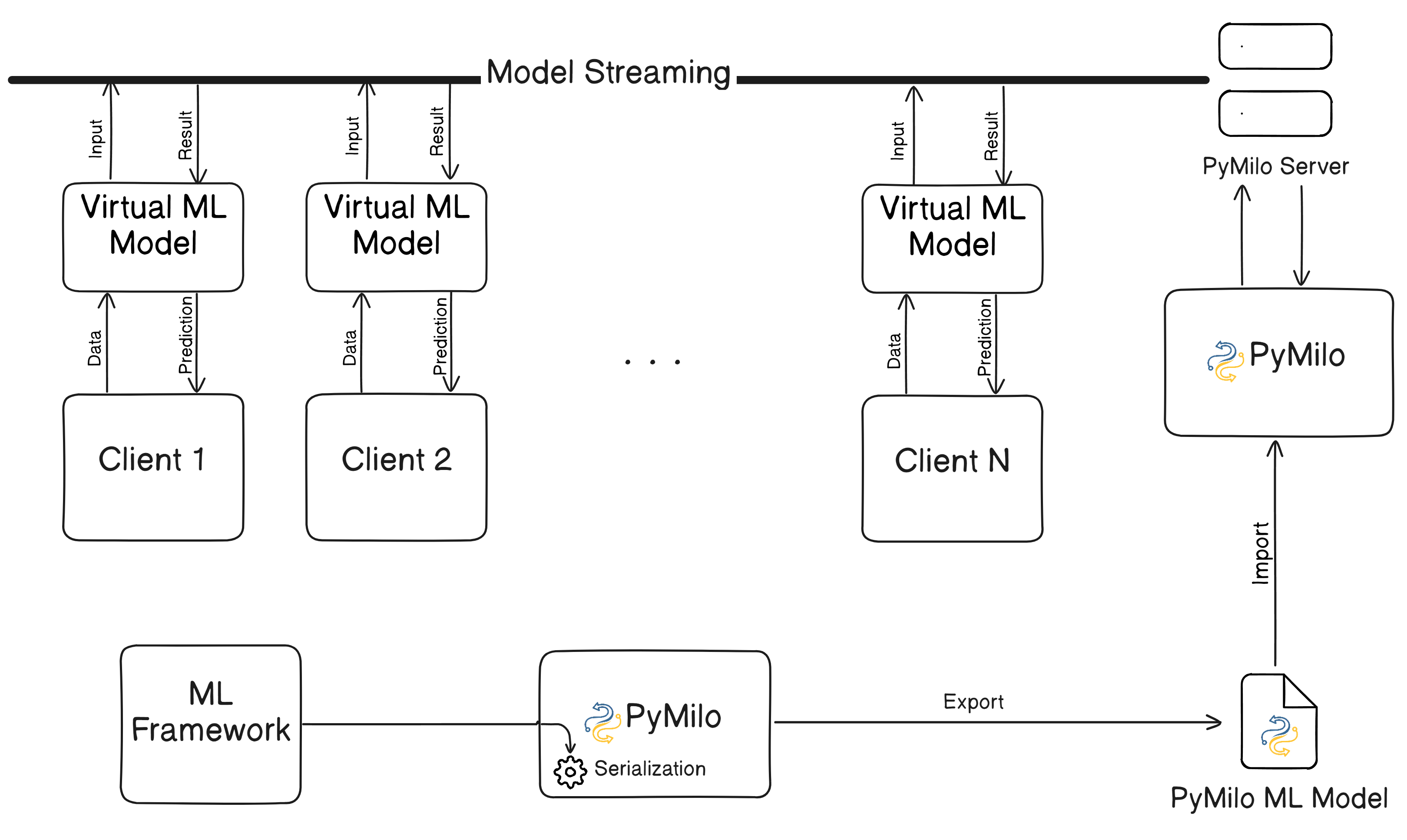}
    \caption{ML Streaming enables the integration of ML models into web services through a server/client architecture. Users can interact with PyMilo-exported models hosted on a PyMiloServer using a PyMiloClient instance.}
    \label{fig:ml-streaming-flow}
    \vspace{0.5cm}
\end{figure*}

This streaming feature also lays the groundwork for the creation of a marketplace-style platform, similar to DockerHub, where pre-trained models can be hosted and shared. By facilitating peer-to-peer (P2P) model sharing, ``ML Streaming" removes the need for centralized storage, allowing users to access and utilize models directly from the hosting node, encouraging a decentralized and more accessible ecosystem for machine learning.

Additionally, ``ML Streaming" positions PyMilo as a key component in Internet of Things (IoT) systems. Through the lightweight ``PyMilo Client," deployed on edge devices, users can interact with remote ML models in real-time. This enables dynamic model updates, predictions, and retraining across distributed edge platforms. Such functionality is particularly beneficial in scenarios where edge computing plays a critical role, such as predictive maintenance, real-time analytics, or adaptive control systems. The decentralized nature of ML Streaming allows models to be deployed on various edge platforms, thereby providing scalable Model as a Service (MaaS) solutions to clients.

\begin{figure*}[!h]
    \centering
    \includegraphics[width=0.7\textwidth]{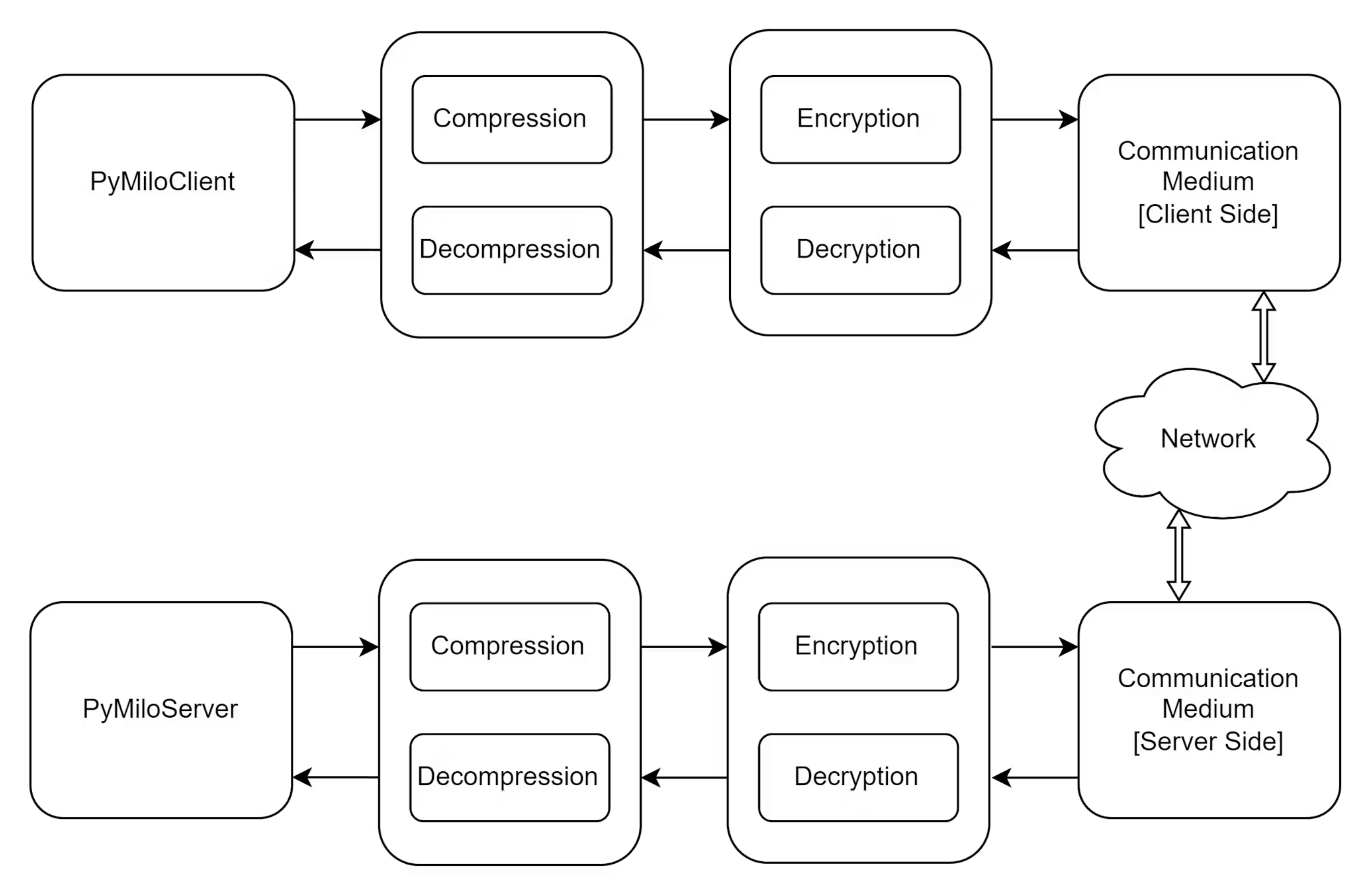}
    \caption{PyMilo's ML Streaming architecture ensures secure and efficient communication between clients and servers. Data transmitted from the PyMiloClient is compressed and encrypted for secure communication, and then decrypted and decompressed upon arrival at the PyMiloServer for processing. The server's response is subsequently compressed and encrypted before being transmitted back to the client, ensuring end-to-end security.}
    \label{fig:ml-streaming-arch}
\end{figure*}
PyMilo is a cross-platform tool that can be executed on various operating systems, that are MacOS ($\geq$10.9), Ubuntu ($\geq$16.04), and Windows ($\geq$8.0). The minimum required Python version for running PyMilo is Python 3.6. A list of PyMilo's dependencies is provided in Table~\ref{tab:dependencies}, which outlines the specific libraries and their corresponding version requirements.
\begin{table}[h]
\centering
\caption{PyMilo's Dependencies}
\label{tab:dependencies}
\begin{tabular}{|l|l|}
\hline
\textbf{PyMilo Base} & \textbf{PyMilo ML Streaming} \\
\hline
numPy ($\geq$1.9.0)~\cite{harris2020array} & uvicorn ($\geq$0.14.0)~\cite{uvicorn014} \\
scikit-learn ($\geq$0.22.2)~\cite{scikit-learn} & fastapi ($\geq$0.68.0)~\cite{fastapi01152} \\
sciPy ($\geq$0.19.1)~\cite{2020SciPy-NMeth} & pydantic ($\geq$1.5.0)~\cite{colvin2023pydantic} \\
requests ($\geq$2.0.0)~\cite{chandra2015python} &  \\
\hline
\end{tabular}
\end{table}
In the following section, we go through a quick demonstration of how PyMilo works.
\section{Demonstration of functionality}\label{sec: functionallity}

In this section, we train a \texttt{LinearRegression} model representing the equation: \(y = x_0 + 2x_1 + 3\). There is a set of data points (\(\mathbf{X}\), \(y\)) available and the model is trained as follows.
\begin{python}
import numpy as np
from sklearn.linear_model import LinearRegression
X = np.array([[1, 1], [1, 2], [2, 2], [2, 3]])
y = np.dot(X, np.array([1, 2])) + 3
# y = 1 * x_0 + 2 * x_1 + 3
model = LinearRegression().fit(X, y)
pred = model.predict(np.array([[3, 5]]))
# pred = [16.] (=1 * 3 + 2 * 5 + 3)
\end{python}
Using PyMilo \texttt{Export} class, the model is serialized and exported into a JSON file.
\begin{python}
from pymilo import Export
Export(model).save("model.json")
\end{python}
As it is shown in Figure~\ref{model_inside}, the exported file is in JSON format and is not an executable. Thus, there is no risk of malware execution. The exported file is transparent, human-readable, and safe. The exported model can easily be opened in an editor. 

\begin{figure}[htb]  % Place the minted code in a figure environment
\centering
\begin{python}
{
  "data": {
    "fit_intercept": true,
    "copy_X": true,
    "n_jobs": null,
    "positive": false,
    "n_features_in_": 2,
    "coef_": {
      "pymiloed-ndarray-list": [
        1.0,
        1.9999999999999993
      ],
      "pymiloed-ndarray-dtype": "float64",
      "pymiloed-ndarray-shape": [
        2
      ],
      "pymiloed-data-structure": "numpy.ndarray"
    },
    "rank_": 2,
    "singular_": {
      "pymiloed-ndarray-list": [
        1.618033988749895,
        0.6180339887498948
      ],
      "pymiloed-ndarray-dtype": "float64",
      "pymiloed-ndarray-shape": [
        2
      ],
      "pymiloed-data-structure": "numpy.ndarray"
    },
    "intercept_": {
      "value": 3.0000000000000018,
      "np-type": "numpy.float64"
    }
  },
  "sklearn_version": "1.5.2",
  "pymilo_version": "1.1",
  "model_type": "LinearRegression"
}
\end{python}
\vspace{-5pt}
\caption{Sample contents of a model exported in a \texttt{model.json} file, illustrating the structure of the model.} 
\label{model_inside}
\end{figure}
Having a non-executable and transparent export file becomes essential when the trained model is provided by an untrustworthy third-party. You can see all the learned parameters of the model in the file and change them if you want. This JSON representation is a transparent version of the trained model.

Using PyMilo's \texttt{Import} class, the model can be loaded back and retrieve the exact same original machine learning model. The imported object is an instance of the \texttt{LinearRegression} class, with identical functionalities and parameters to the original one.
\begin{python}
from pymilo import Import
model = Import("model.json").to_model()
pred = model.predict(np.array([[3, 5]]))
# pred = [16.] (=1 * 3 + 2 * 5 + 3)
\end{python}
\section{Quality Control}
To ensure the reliability and accuracy of PyMilo, we employed a comprehensive testing framework using the \verb|pytest| unit test framework~\cite{pytest4.3}. Currently, PyMilo supports Scikit-learn, with plans to expand support to PyTorch and Tensorflow.\\
Our testing framework covers 100 distinct machine learning models available in Scikit-learn. Each model undergoes a systematic evaluation within the pipeline, which involves: (1) fitting the model on a designated training dataset, (2) computing the evaluation metrics—Mean Squared Error (MSE) and coefficient of determination ($R^2$) for regression models, hinge loss and accuracy score for classification models, and (3) exporting the model using PyMilo and storing it in JSON format.

To evaluate the integrity of the transportation process, we subsequently retrieve the model using the PyMilo Import class. After transportation, we recalculate the corresponding evaluation metrics (MSE and $R^2$ for regression models, hinge loss and accuracy score for classification models). For clustering models, we compare the predictions on the same test dataset before and after transportation. The criterion for passing a test case is established such that the cumulative absolute differences between the pre-PyMilo and post-PyMilo evaluation metrics, or the differences in clustering predictions, must be less than \(10^{-8}\). This pipeline is executed across eight Python versions (from Python 3.6 to Python 3.13) and three distinct operating systems (Linux, macOS, and Windows), resulting in a total of 24 testing stages. Each Pull Request (PR) should pass all of the automated tests and be approved by at least 2 main reviewers in order to get approved and be merged.
\section{Applications and Reuse potential}
\subsection{Use Cases}
PyMilo is designed to be highly reusable, and we encourage researchers and practitioners in machine learning and MLOps to leverage its capabilities. The software can be easily cloned and modified to suit specific use cases, allowing users to tailor the library to their needs.
PyMilo focuses on a gap in the AI ecosystem, particularly for scenarios that prioritize safety, transparency, and authenticity: 
\begin{itemize} 
\item Model Sharing: PyMilo enables the secure exchange of machine learning models. Its transparency and human-readability ensure that the models remain unchanged, safe, and harmless throughout the sharing process. 
\item End-to-end Deployment: With PyMilo, models can be exported after the development phase and be imported and deployed on production. This simplifies the transition from development to production, ensuring consistent performance.
\item Safety and Authenticity: PyMilo provides traceability and verification features that confirm the model's origin and prevent unauthorized alterations. This ensures that models used in critical industries, such as healthcare, meet safety and regulatory requirements. \end{itemize}
\subsection{Modification and Extension}
Users can modify PyMilo's transporters to suit their specific needs, and add new functionalities during the serialization or deserialization process. We encourage contributors to fork the repository and propose changes as pull requests to the main library base. This allows users to create tailored versions of the library for their specific use cases, while also benefiting the broader community.
\subsection{Support Mechanism}
We provide comprehensive documentation in the GitHub repository page, including instructions for installation and usage of the PyMilo library. We also actively monitor the GitHub issues and provide a Discord channel for users to ask questions, discuss issues, and request new features.
\section{Limitations and future works}
Currently, PyMilo supports all machine learning models available in the scikit-learn library. Our next goal is to extend support to PyTorch models, followed by TensorFlow. Additionally, for the ML Streaming feature, we plan to expand protocol compatibility by adding support for WebSocket and MQTT, making PyMilo a more versatile and extensible tool.
\section*{Acknowledgements}
We sincerely appreciate \href{https://www.python.org/psf/}{Python Software Foundation (PSF)} that grants PyMilo library partially for versions \textbf{1.0 and 1.1}. The \href{https://www.python.org/psf/}{PSF} is the organization behind Python. Their mission is to promote, protect, and advance the Python programming language and to support and facilitate the growth of a diverse and international community of Python programmers.

We would like to express our gratitude to \href{https://trelis.com/}{Trelis Research} for partially funding PyMilo's version \textbf{1.0}. The \href{https://trelis.com/}{Trelis Research} provides tools and tutorials for businesses and developers looking to fine-tune and deploy large language models.

We also thank Zahra Mobasher Amini for designing the PyMilo's logo.

\bibliographystyle{IEEEtran}  
\bibliography{bibfile} 
\vspace{12pt}

\end{document}